\pdfoutput=1

\documentclass[11pt]{article}

\usepackage[]{acl}

\usepackage{times}
\usepackage{latexsym}

\usepackage[T1]{fontenc}

\usepackage[utf8]{inputenc}

\usepackage{microtype}

\usepackage{amsmath}
\usepackage{amsfonts}
\usepackage{bm}
\usepackage{booktabs}
\usepackage{multirow}
\usepackage{CJKutf8}
\usepackage{pgfplots}
\usetikzlibrary{patterns}
\definecolor{myred}{HTML}{FA7F6F}
\definecolor{myblue}{HTML}{82B0D2}

\newcommand{\tabref}[1]{\tablename~\ref{table:#1}}
\newcommand{\figref}[1]{\figurename~\ref{figure:#1}}

\title{Domain-Specific NER via Retrieving Correlated Samples}

\author{
Xin Zhang$^1$, Yong Jiang, Xiaobin Wang, Xuming Hu$^2$, \\
\textbf{Yueheng Sun$^3$, Pengjun Xie, Meishan Zhang}$^{4}$\thanks{~~Corresponding author.}\\
$^1$School of New Media and Communication, Tianjin University, China\\
$^2$School of Software, Tsinghua University, China\\
$^3$College of Intelligence and Computing, Tianjin University, China\\
$^4$Institute of Computing and Intelligence, Harbin Institute of Technology (Shenzhen), China\\
\texttt{hsinz@tju.edu.cn, hxm19@mails.tsinghua.edu.cn} \\
\texttt{jiangyong.ml@gmail.com, czwangxiaobin@foxmail.com} \\
\texttt{yhs@tju.edu.cn, xpjandy@gmail.com, zhangmeishan@hit.edu.cn} \\
}

\pgfplotsset{compat=1.17}

\begin{document}
\begin{CJK}{UTF8}{gbsn}

\maketitle
\begin{abstract}
Successful Machine Learning based Named Entity Recognition models could fail on texts from some special domains, for instance, Chinese addresses and e-commerce titles, where requires adequate background knowledge.
Such texts are also difficult for human annotators.
In fact, we can obtain some potentially helpful information from correlated texts, which have some common entities, to help the text understanding.
Then, one can easily reason out the correct answer by referencing correlated samples.
In this paper, we suggest enhancing NER models with correlated samples.
We draw correlated samples by the sparse BM25 retriever from large-scale in-domain unlabeled data.
To explicitly simulate the human reasoning process, we perform a training-free entity type calibrating by majority voting.
To capture correlation features in the training stage, we suggest to model correlated samples by the transformer-based multi-instance cross-encoder.
Empirical results on datasets of the above two domains show the efficacy of our methods.
\end{abstract}

\section{Introduction}

Named Entity Recognition (NER), which first locates entity positions and then labels their types sequentially, is a fundamental topic in both academia and industry \cite{DBLP:journals/tkde/LiSHL22}.
Normal NER models consider the input samples to be independent of each other, learning the common intra-instance patterns and making predictions in a sequential way.
This paradigm has shown surprising successes in decades, especially with the help of emerging deep learning
\cite{shang-etal-2018-learning,DBLP:conf/aaai/0001FLH18,liu-etal-2019-towards,DBLP:conf/aaai/LuoXZ20,lison-etal-2020-named,fang-etal-2021-tebner,meng-etal-2021-distantly}.

\begin{figure}
\centering \includegraphics[scale=0.33]{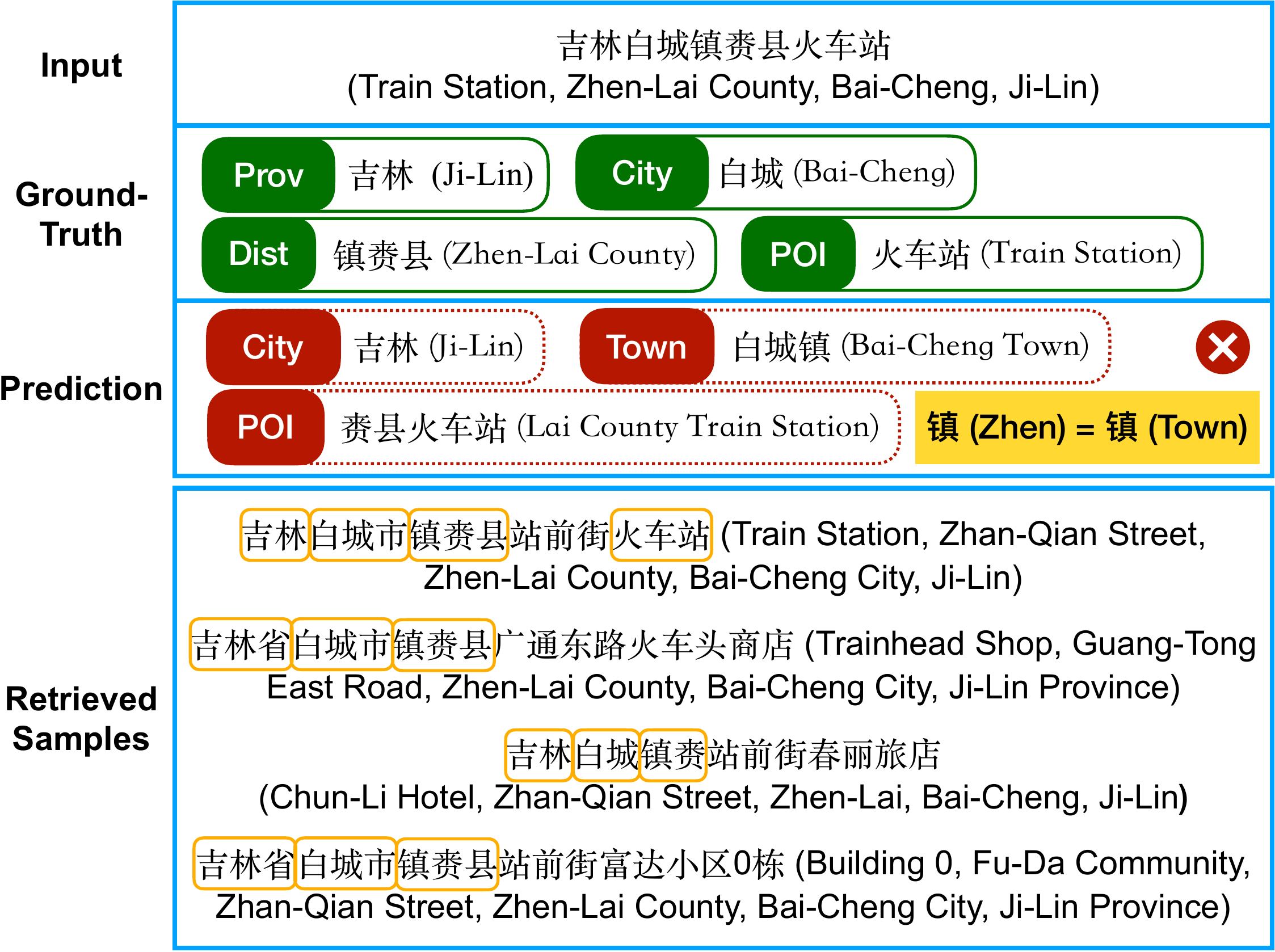}
\caption{An address example with retrieved texts.
The model incorrectly predicted ``白城镇 (Baicheng Town)'' and ``赉县火车站 (Lai County Train Station)'' because they match common patterns, i.e. ``XX镇'' (Xx Town) and ``X县火车站'' (X County Train Station). ``吉林'' is ambiguous, which is both a province and a city.
}
\label{figure:example}
\end{figure}

However, learned models will fail at some hard cases, which would be inevitably encountered in real scenarios \cite{li-etal-2019-neural-chinese,ding-etal-2019-neural}.
\figref{example} shows an example of the Chinese address domain.
This kind of bad cases can not be easily solved by annotating more relevant training data\footnote{
Because this pattern is indeed correct in most cases. This problem also exists in models with internal larger datasets.
}.
For human annotators, this case is ambiguous as well if no extra information is given,
for instance, we can not distinguish the type of ``吉林 (Jilin)'' without affixes ``省 (Province)'' or ``市 (City)''.
This demonstrates that obtaining background knowledge and information is crucial to the text understanding.

\begin{figure*}
\centering \small
\includegraphics[scale=0.48]{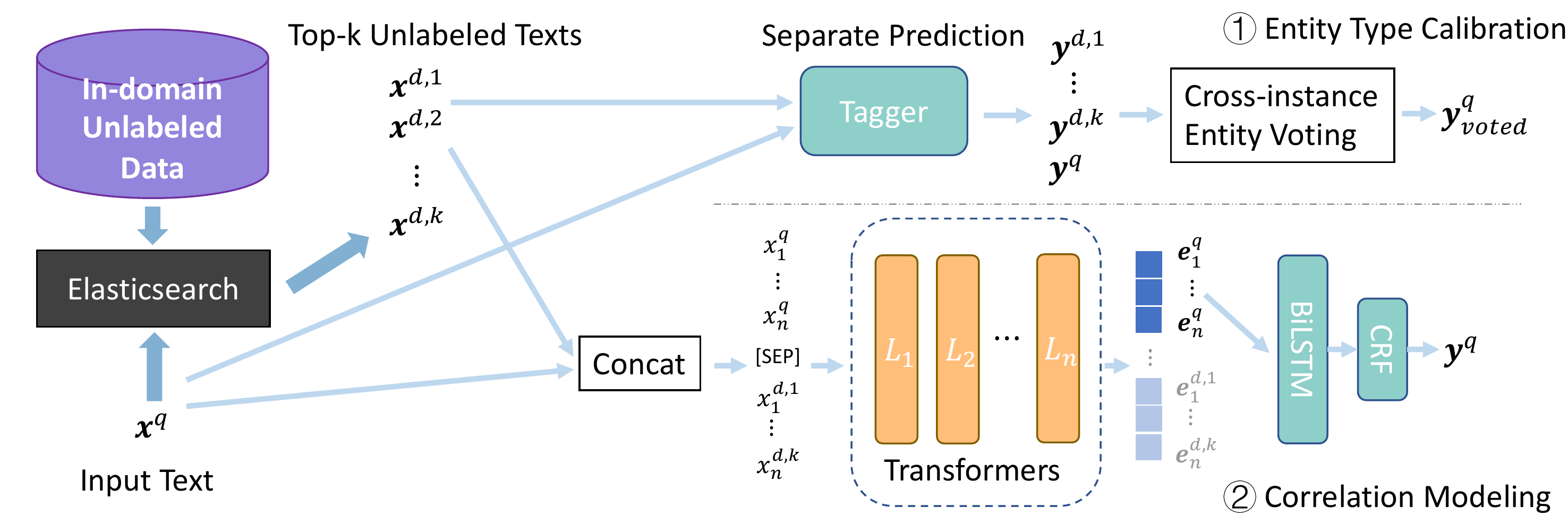}
\caption{
The overview of our suggested methods.
}
\label{figure:method}
\end{figure*}
\setlength{\textfloatsep}{5pt plus 2pt minus 2pt}

Learning from correlated or nested data is mainly studied in Machine Learning and Computer Vision \cite{DBLP:conf/ijcai/DundarKBR07,DBLP:conf/ijcai/ChoiW19,DBLP:conf/icml/ChoiLW21}.
Images in sub-groups naturally show a high degree of correlation on both features and labels, and come with nested structures \cite{DBLP:conf/ijcai/DundarKBR07,DBLP:conf/ijcai/ChoiW19}, such as different regions of interest could be drawn from the same objects.
In the address and e-commerce domain, texts are also highly correlated in nature.
For example, two addresses may belong to the same city or refer to the same location,
e-commerce product titles could come from the same brand, or they are just the same product.
In \figref{example}, with correlated texts, annotators can infer that the ``吉林 (Jilin)'' and ``白城 (Baicheng)'' are short forms of ``吉林省 (Jilin Province)'' and ``白城市 (Baicheng City)''.
Hence, we argue that correlated samples could offer sufficient disambiguation information for NER models as well.
Such kind of inductive bias is seldom considered in previous NLP studies.

In this work, we propose to enhance NER models by 
modeling and inferencing with the correlated samples.
We first draw the correlated samples from in-domain large-scale unlabeled data by the retrieval engine \cite{elasticsearch}.\footnote{
Recently, \citet{wang-etal-2021-improving} and \citet{DBLP:journals/corr/abs-2202-09022} studied retrieving external contexts from Google or Baidu for standard NER datasets, which is quite different from our idea of modeling correlated samples for specific domains.
}
Then, we suggest two methods:
(1) we perform an entity type calibrating by parallelly predicting the input text and all retrieved samples by the off-the-shelf NER model, and then aggregating the final labels by majority voting;
(2) we propose to model the correlations by transformers via multi-instance cross-encoders to enhance the NER feature vectors.

To evaluate our methods, we conduct experiments on two open-access datasets \cite{ccks2021track2,ding-etal-2019-neural} of the aforementioned two domains.
We implement our methods based on a strong BiLSTM-CRF model with NEZHA \cite{DBLP:journals/corr/abs-1909-00204} representation.
Empirical results show that our methods outperform all baselines, and achieve promising results in the simulated low-resource setting.
Finally, we present several analyses to understand our methods comprehensively.

\section{Approach}

\subsection{Unlabeled Data Retrieval}\label{section:retriever}
In the address (resp. e-commerce) domain, some texts naturally possess entity co-reference relations, for instance, they may belong to the same city (resp. brand) or represent the same location (resp. product).
We call such texts, which usually have entities with the same semantic but different expressions, {\bf correlated samples}.
Since these texts are highly structured and of limited vocabulary, showing a high degree of lexical overlap.
We could draw correlated samples for a given text by taking it as a query and retrieving the domain-specific database with text similarity measurements.

We implement an efficient BM25 \cite{DBLP:conf/sigir/RobertsonW94} retriever by an off-the-shelf retrieval engine \cite{elasticsearch}.
For a cleaned large-scale in-domain unlabeled corpus, we create the Elasticsearch index by the build-in standard analyzer.
Then, we can retrieve top-K samples by BM25 scores of an input text in nearly real-time.

\subsection{Entity Type Calibrating}\label{section:vote}
As shown in \figref{example}, correlated sample can help the entity disambiguation.
If this kind of entity appears in correlated samples, human annotators can decide its type by referring to answers of correlated samples.
For NER models, we suggest achieving this process by entity-level (or span-level) majority voting.
Concretely, we first use a model (e.g., baseline) to extract entities of the input text and each correlated sample parallelly, and then re-assign labels of shared entities by majority voting.

\subsection{Correlation Modeling}\label{section:concat}
To further capture sample correlations in the training time, we suggest modeling correlated samples by the cross-encoder~\cite{reimers-gurevych-2019-sentence}, letting transformers learn complex correlation patterns among samples.
Specifically, we concatenate the input text with retrieved samples by the separator (i.e., [SEP]), and then encode them by pretrained language models.
Finally, only the contextual embeddings of the input text are fed into the NER tagger (here BiLSTM-CRF).
With this simple strategy, NER models could benefit from the contrastive view between multiple correlated samples and understand the query instance better.

\section{Experiments}

\begin{table}
\setlength{\tabcolsep}{3pt}
\centering
\resizebox{0.48\textwidth}{!}{
\begin{tabular}{l|ccc|ccc}
\toprule
 & \multicolumn{3}{c|}{Micro} & \multicolumn{3}{c}{Macro} \\
Method & P & R & F1 & P & R & F1 \\
\midrule
\multicolumn{7}{c}{Chinese Address} \\
\midrule
Human & 93.04 & 92.01 & 92.52 & 87.83 & 84.52 & 86.14 \\
\midrule
BC & 85.56 & 83.90 & 84.72 & 82.20 & 76.48 & 79.24 \\
NEZHA-BC & 91.29 & 90.62 & 90.95 & 86.41 & 84.68 & 85.53 \\
\midrule
Entity-Voting$^\dagger$ & 91.67 & 91.00 & 91.34 & 86.70 & 84.83 & 85.76 \\
Cross-Encoder$^\dagger$ & \bf 92.41 & \bf 91.95 & \bf 92.18 & \bf 87.25 & \bf 85.71 & \bf 86.48 \\
\midrule
Self-Training & 91.57 & 91.02 & 91.29  & 86.65 & 85.37 & 86.01 \\
Biaffine      & 91.35 & 90.25 & 90.80  & 86.32 & 84.59 & 85.45 \\
Seq2set       & 89.43 & 87.69 & 88.55  & 83.89 & 80.12 & 81.96 \\
Locate\&Label & 90.28 & 87.76 & 89.00  & 85.95 & 82.29 & 84.08 \\
PIQN          & 90.27 & 87.83 & 89.03  & 86.04 & 80.28 & 83.06 \\

\midrule
\multicolumn{7}{c}{E-commerce} \\
\midrule
BC & 65.31 & 62.54 & 63.90 & 58.88 & 50.38 & 54.30 \\
NEZHA-BC & 82.73 & 83.23 & 82.98 & 79.35 & 78.04 & 78.69 \\
\midrule
Entity-Voting$^\dagger$ & 82.83 & 83.33 & 83.08 & 79.56 & 78.19 & 78.87 \\
Cross-Encoder$^\dagger$ & \bf 83.49 & 83.74 & \bf 83.61 & \bf 81.45 & \bf 79.34 & \bf 80.38 \\
\midrule
Self-Training & 81.51 & \bf 85.25 & 83.34 & 78.89 & 79.29 & 79.09 \\
Biaffine      & 81.91 & 84.06 & 82.97  & 80.14 & 79.05 & 79.59\\
Seq2set       & 82.77 & 81.65 & 82.21  & 81.39 & 76.44 & 78.84 \\
Locate\&Label & 80.43 & 83.21 & 81.80  & 76.63 & 78.22 & 77.42 \\
PIQN          & 83.43 & 82.54 & 82.98  & 81.23 & 75.60 & 78.31 \\
\midrule
BERT-CLS \citeyearpar{devlin-etal-2019-bert} & 77.06 & 80.65 & 78.81 &-&-&- \\
MRC-NER \citeyearpar{li-etal-2020-unified} & 79.47 & 78.30 & 78.88 &-&-&- \\
CoFEE-BERT \citeyearpar{mengge-etal-2020-coarse} & 79.13 & 80.34 & 79.73 &-&-&- \\
CoFEE-MRC \citeyearpar{mengge-etal-2020-coarse} & 80.26 & 78.88 & 79.56 &-&-&- \\
\bottomrule
\end{tabular}
}
\caption{
Main results. $^\dagger$ means statistically significant.
}\label{table:result}
\end{table}

\subsection{Settings}

\paragraph{Datasets.}
For the {\bf Chinese Address} domain, we use the recently published dataset from CCKS competition \cite{ccks2021track2}. 
It is annotated by 21 classes of address elements and contains 8856, 1970, 4000 addresses for train, dev, and test sets.
For the {\bf E-commerce} domain, we use the dataset released by \citet{ding-etal-2019-neural}. 
It is collected from e-commerce product titles and annotated by PROD (product) and BRAN (brand) types. It has 3983, 499, 498 sentences\footnote{
We remove a few sentences that are particularly long and do not contain entities.
} for train, dev, and test sets.
For our \textbf{retrieval-based} methods, we process and index our internal in-domain unlabeled data with Elasticsearch, obtaining 400M and 600M samples for the address and e-commerce domain, respectively.

\paragraph{Evaluation.}
We employ entity-level exact precision, recall, and F1-measure and report both micro and macro aggregations.
All experiments of the same setting are conducted by 8 different random seeds.
We test the best model of the devset, and the average scores are reported.
We regard a result as statistically significant when the p-value is below 0.05 by the paired t-test with baseline NEZHA-BC.

\paragraph{Implementation.}
We choose the BiLSTM-CRF \cite{lample-etal-2016-neural} to achieve NER task,
and use NEZHA-base \cite{DBLP:journals/corr/abs-1909-00204} as the embedding module.
The BiLSTM hidden size is set to 384 for each direction.
We apply the dropout \cite{DBLP:journals/jmlr/SrivastavaHKSS14} with probabilities 0.5 and 0.2 to NEZHA embeddings for address and e-commerce, and 0.2 to BiLSTM features.
We set the batch size to 32 and use the AdamW \cite{DBLP:journals/corr/abs-1711-05101} optimizer with a constant lr 1e-3 and 1e-5 to update BiLSTM-CRF and NEZHA parameters.

For the entity type calibrating, we use the top 100 and 50 retrieved samples for address and e-commerce, respectively.
For the correlating modeling, we limit the max sample number to 12 and the max sequence length to 256.

\paragraph{Baselines.}
We denote the BiLSTM-CRF with random character embedding (resp. NEZHA) by \textbf{BC} (resp. \textbf{NEZHA-BC}).
We implement several state-of-the-art methods, i.e., \textbf{Biaffine} \cite{yu-etal-2020-named}, \textbf{Seq2set} \cite{DBLP:conf/ijcai/Tan0Z0Z21}, \textbf{Locate\&Label} \cite{shen-etal-2021-locate}, \textbf{PIQN} \cite{shen-etal-2022-piqn}.
We also implement {\bf Self-Training} based on NEZHA-BC and the unlabeled data of the same size as our cross-encoder.
We include e-commerce results from \citet{mengge-etal-2020-coarse} for comparison.

\subsection{Main Results}\label{section:result}

\begin{figure}\centering
\includegraphics[scale=0.8]{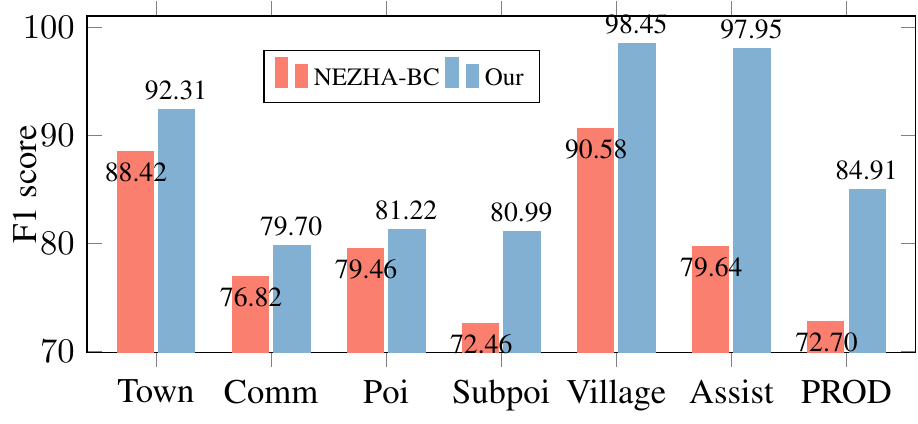}
\caption{
Test F1 of some hard entity types from main results, where their scores are less than the overall value.
}
\label{figure:type}
\end{figure}

As shown in \tabref{result}, our training-free calibrating method consistently outperforms our implemented baselines on both datasets, which verifies our intuition that modeling the correlation between samples is important in processing domain-specific texts.
By leveraging the retrieved samples in the training stage (Cross-Encoder), our approaches gain a significant performance boost.
This indicates that these retrieved samples not only provide extended entity information (such as 白城$\longrightarrow$白城市), but also supply sufficient disambiguate signals for entity understanding (such as 镇赉县火车站 v.s 镇赉县站前街火车站).
When compared with other recent state-of-the-art NER methods (Biaffine, Seq2set, Locate\&Label, and PIQN), our approaches outperform them by a large margin.
It is worth noting that our model outperforms the self-training (whose unlabeled corpus is in the same scale of samples we modeled), demonstrating that the correlation modeling is more effective.
Then we plot detailed scores by categories whose F1 score is less than the overall F1 in \figref{type}.
All of these difficult categories are significantly improved, showing that the correlated samples are helpful.

\begin{table}
\setlength{\tabcolsep}{3pt}
\centering
\resizebox{0.48\textwidth}{!}{
\begin{tabular}{cc|cccccc}
\toprule
\multicolumn{2}{c|}{Method} & 100\% & 50\% & 20\% & 10\% & 5\% & 3\% \\
\midrule
\multicolumn{8}{c}{Chinese Address} \\
\midrule
\multirow{3}{*}{Micro} & NEZHA-BC &
90.95 & 90.04 & 88.79 & 87.57 & 86.56 & 84.69 \\
\cmidrule{2-8}
& Cross- & 
92.18 & 91.56 & 90.33 & 89.13 & 88.61 & 86.82 \\
& encoder & $\uparrow$1.23 & $\uparrow$1.52 & $\uparrow$1.54 & $\uparrow$1.56 & $\uparrow$2.05 & $\uparrow$2.13 \\
\midrule
\multirow{3}{*}{Macro} & NEZHA-BC &
85.53 & 84.22 & 82.14 & 78.08 & 75.99 & 72.78 \\
\cmidrule{2-8}
& Cross- &
86.48 & 85.36 & 83.43 & 79.66 & 78.31 & 75.41 \\
& encoder & $\uparrow$0.95 & $\uparrow$1.14 & $\uparrow$1.29 & $\uparrow$1.58 & $\uparrow$2.32 & $\uparrow$2.63 \\

\midrule
\multicolumn{8}{c}{E-commerce} \\
\midrule
\multirow{3}{*}{Micro} & NEZHA-BC &
82.98 & 81.54 & 79.52 & 77.89 & 75.80 & 73.86 \\
\cmidrule{2-8}
& Cross- &
83.61 & 82.21 & 80.29 & 78.99 & 77.18 & 74.60 \\
& encoder & $\uparrow$0.63 & $\uparrow$0.67 & $\uparrow$0.77 & $\uparrow$1.10 & $\uparrow$1.38 & $\uparrow$0.74 \\
\midrule
\multirow{3}{*}{Macro} & NEZHA-BC &
78.69 & 77.03 & 75.17 & 72.86 & 69.63 & 67.11 \\
\cmidrule{2-8}
& Cross- &
80.21 & 78.21 & 76.47 & 74.28 & 71.47 & 68.03 \\
& encoder & $\uparrow$1.52 & $\uparrow$1.18 & $\uparrow$1.30 & $\uparrow$1.42 & $\uparrow$1.84 & $\uparrow$0.92 \\
\bottomrule
\end{tabular}
}
\caption{
Test F1 scores at various low-resource settings.
}\label{table:low}
\end{table}

\paragraph{Other Results.}
For the Chinese Address dataset, we also report the performance of human annotators without extra information provided.
Notably, our approaches achieve comparable performance with humans,
which empirically verifies that modeling text correlation with the retrieval perspective might have the possibility to simulate human expert annotations.
For the E-commerce dataset, we also report other published results.
Our NEZHA-BC is comparable with all the baseline implementations.

\begin{table}
\setlength{\tabcolsep}{3pt}
\centering
\resizebox{0.45\textwidth}{!}{
\begin{tabular}{c|cccccc}
\toprule
\#Address & 400M & 40M & 10M & 4M & 400k & 100k \\
\midrule
Micro F1 & 92.18 & 92.07 & 91.89 & 91.53 & 91.31 & 91.16 \\
\bottomrule
\end{tabular}
}
\caption{
Test F1 scores of our Cross-encoder in various sizes of unlabeled data for retrieval in address domain.
}\label{table:diffun}
\end{table}

\begin{figure}\centering
\includegraphics[scale=0.8]{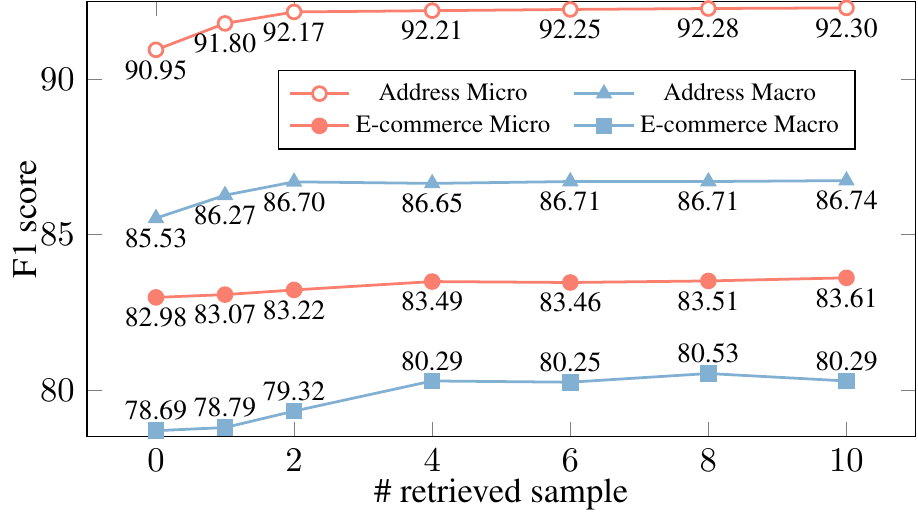}
\caption{
Test F1 scores of incooperating different correlated sample num by our cross-encoder.
}\label{figure:num}
\end{figure}

\subsection{Analysis}\label{section:analysis}
We conduct fine-grained analyses of cross-encoder.

\paragraph{Different Sizes of Labeled Data.}
Our idea essentially introduces extra in-domain data to the predictive models.
Hence we can suppose that our methods will achieve larger improvements in the low-resource scenario.
To verify this, we train the baseline and our cross-encoder in simulated smaller trainsets, which are sampled from the original trainset by different proportions.
\tabref{low} demonstrates the test f1 scores of these two models in different settings.
We can roughly say that the score difference increases as the sampling ratio decreases, which is in line with our intuition.

\paragraph{Different Sizes of Correlated Samples.}
In the above experiments, we limit the max sequence length of our cross-encoder to 256 for efficiency.
Here we relax this constraint to investigate the influence of encoded sample num (from 0\footnote{
The 0 samples cross-encoder degrade to the baseline NEZHA-BC.
} to top 10 retrieved texts) in cross-encoder on both two domains.
As shown in \figref{num}, the performance increment is significant at the lower sample number.
And adding more relatively low-ranking samples is of limited gains.

\paragraph{Different Sizes of Unlabeled Data.}
All of the previous experiments are based on the same large-scale in-domain unlabeled data, which almost reach the billion-level (400M and 600M samples for address and e-commerce, respectively).
We also sample several smaller unlabeled corpus (i.e., 40M, 10M, 4M, 400k, 100k) and re-train our cross-encoder.
As shown in Table \ref{table:diffun}, with the size of unlabeled data declines, the retrieved samples are less relevant, the improvements of our model are lower.
Interestingly, this experiments also could reflect the effect of \textbf{unlabeled data quality} to the performance of our cross-encoder.
The higher the quality of the data, the more correlated samples can be retrieved.
The behavior of low-quality unlabeled data is similar to the small size data.

\begin{table}
\setlength{\tabcolsep}{3pt}
\centering
\resizebox{0.45\textwidth}{!}{
\begin{tabular}{c|ccc}
\toprule
Method & NEZHA-BC & Entity-Voting & Cross-Encoder \\
\midrule
Seconds & 14.73 & 500+ & 38.41 \\
\bottomrule
\end{tabular}
}
\caption{
Running times of different methods on the address domain testset, which has 4,000 texts.
}\label{table:speed}
\end{table}

\paragraph{Running Speed of Different Methods.}
Another key concern of our methods is the running speed.
The entity-voting needs parallelly decode dozens of texts, and the cross-encoder will significantly enlarge the text length.
We measured the running time of several methods on the testset of the address domain dataset.
As demonstrated in Table \ref{table:speed}, the entity-voting is truely slower than other methods in an order of magnitude.
But the cross-encoder just took about twice as long as the baseline NEZHA-BC.
This is because the most time-consuming part is the CRF, where the concatenated samples are droped before the CRF.
So it can avoid the redundant decoding in the entity-voting, and has a higher running speed.
Besides, the forward of pretrained language models are highly optimized.

\begin{table}
\centering
\resizebox{0.40\textwidth}{!}{
\begin{tabular}{c|ccc}
\toprule
Method & P & R & F1 \\
\midrule
NEZHA-BC & 93.02 & 92.55 & 92.78 \\
+Cross-Encoder & 94.12 & 93.67 & 93.89 \\
\bottomrule
\end{tabular}
}
\caption{
Performance on the devset of the address dataset.
NEZHA-BC is the baseline NEZHA-BiLSTM-CRF.
}\label{table:dev}
\end{table}

\paragraph{Address dataset Development Performance.}
Since the testset labels are not released \cite{ccks2021track2}, we report devset scores for comparison.
As shown in Table \ref{table:dev}, it is consistent with our main results.

\subsection{Discussion}
Retrieval-augmented models are showing state-of-the-art performance in many NLP tasks, such as Dialogue \cite{weston-etal-2018-retrieve}, Neural Machine Translation \cite{zhang-etal-2018-guiding}, Question Answering \cite{izacard-grave-2021-leveraging}, and Language Modeling \cite{DBLP:conf/icml/GuuLTPC20,DBLP:conf/icml/YaoZYY22,DBLP:conf/icml/BorgeaudMHCRM0L22}.
Our work aims to model the internal correlation within sub-groups of samples.
We first retrieve correlated sample groups for a given input by the off-the-shelf Elasticsearch engine.
Then, we propose painlessly calibrating entity type and transformer-based correlation modeling, where the latter one is similar to \citet{wang-etal-2021-improving}.
Our recent work \cite{wang-etal-2022-damo} also investigated retrieving knowledge from the Wikipedia, which can augment the context of NER inputs and shows significant improvements in SemEval-2022 Task 11 Multilingual NER.

This work could be further investigated with some more sophisticated techniques, such as example-based learning \cite{gao-etal-2021-making,lee-etal-2022-good,liu-etal-2022-hiure}.
Meanwhile, it also may help the NER task to extend to the low-resource and zero-shot scenarios \cite{meng-etal-2021-distantly,zhang-etal-2021-crowdsourcing,hu-etal-2021-gradient,lu-etal-2022-unified,hu-etal-2020-selfore}.

\section{Conclusion}
In this work, we investigated utilizing naturally correlated samples to improve current NER models on the Chinese address and e-commerce domain.
We propose to retrieve correlated samples for the given text by the BM25 and elasticsearch engine.
To explore the correlations in a light way, we suggest calibrating the predicted entity types by cross-instance entity voting.
To further incorporate these correlated samples into model training, we use multi-instance cross-encoders to learn more complex correlations.
Empirical results show that the painless entity type calibrating improved the performance to some extent, and modeling correlations by cross-encoders achieved the state-of-the-art performance.
We hope this idea could benefit the similar scenario/domains of other tasks.

We will release our code and data at \href{https://github.com/izhx/NER-unlabeled-data-retrieval}{github.com/izhx/NER-unlabeled-data-retrieval} to facilitate future research.

\section*{Acknowledgements}
We thank all reviewers for their hard work.
This research is supported by grants from the National Natural Science Foundation of China (No. 62176180).

\section*{Ethical Statement}
All texts are anonymized.

\bibliography{sigir22}

\end{CJK}
\end{document}